\documentclass[dvipsnames,fleqn,10pt]{wlscirep}
\usepackage[utf8]{inputenc}
\usepackage[T1]{fontenc}

\usepackage{float}
\usepackage{verbatim}
\restylefloat{figure}
\restylefloat{table}
\usepackage{inputenc}
\usepackage[T1]{fontenc}
\usepackage{mathtools}
\usepackage{amssymb}
\usepackage{amsmath}
\usepackage{algorithm}
\usepackage[noend]{algpseudocode}
\usepackage{upquote}
\usepackage{soul,color}
\usepackage{enumitem}

\usepackage{subcaption}
\usepackage{rotating}
\usepackage{multirow}
\usepackage{booktabs}
\usepackage{xurl}
\usepackage{xcolor}
\usepackage[nameinlink]{cleveref}
\title{Feature space reduction method for ultrahigh-dimensional, multiclass data: Random forest-based multiround screening (RFMS)}

\author[1]{Gergely Hancz\'ar}
\author[2]{Marcell Stippinger}
\author[1]{D\'avid Han\'ak}
\author[2,3,*]{Marcell T.\ Kurbucz}
\author[1]{Oliv\'er M. T\"orteli}
\author[1]{\'Agnes Chripk\'o}
\author[2]{Zolt\'an Somogyv\'ari}

\affil[1]{Cursor Insight Ltd., 20-22 Wenlock Road, N17GU London, United Kingdom}
\affil[2]{Department of Computational Sciences, Wigner Research Centre for Physics, 29-33 Konkoly Thege Mikl\'os Street, H-1121 Budapest, Hungary}
\affil[3]{Institute of Data Analytics and Information Systems, Corvinus University of Budapest, 8 F\H{o}v\'am Square, H-1093, Hungary}
\affil[*]{kurbucz.marcell@wigner.hu}


\begin{abstract}
In recent years, numerous screening methods have been published for ultrahigh-dimensional data that contain hundreds of thousands of features; however, most of these features cannot handle data with thousands of classes. Prediction models built to authenticate users based on multichannel biometric data result in this type of problem. In this study, we present a novel method known as \emph{random forest-based multiround screening (RFMS)} that can be effectively applied under such circumstances. The proposed algorithm divides the feature space into small subsets and executes a series of partial model builds. These partial models are used to implement tournament-based sorting and the selection of features based on their importance. To benchmark RFMS, a synthetic biometric feature space generator known as \textit{BiometricBlender} is employed. Based on the results, the RFMS is on par with industry-standard feature screening methods while simultaneously possessing many advantages over these methods.
\end{abstract}

\begin{document}
\setlength{\parskip}{0pt}

\flushbottom
\maketitle
\thispagestyle{empty}

\section*{Introduction}
\label{sec:intro}

\noindent
The understanding of human motor coordination and the building of prediction models to meet various business needs have become widely studied topics in fields such as neurology and cybersecurity. With the help of adequate sensors, gestures, walking, handwriting, eye movement, or any other human motor activity can be transformed into multidimensional time series. However, from a general perspective, any fixed set of features is either uncharacteristic of these time series, or it is too large for resource efficient classification. Thus, instead of computing an \emph{a priori} defined, conveniently small set of features, a promising alternative strategy is to create an ultrahigh-dimensional dataset that consists of hundreds of thousands of features and to search for the most informative minimal subset\cite{wang2009forward}. In this process, as well as in many other machine learning applications, the evaluation of feature importance and the elimination of irrelevant or redundant predictors has become one of the crucial elements in improving the performance of algorithms\cite{tan2022feature}. This elimination can increase the accuracy of the learning process and reduce the resource needs of model building. The statistical challenges of high dimensionality have been thoroughly reviewed in\cite{clarke2008properties,johnstone2009statistical,li2017challenges}.

Traditional variable selection methods do not usually work well in ultrahigh-dimensional data analysis because they aim to specifically select the optimal set of active predictors\cite{ferri1994, ni2017adjusted, borutaalg, speiser2019}. It has also been reported that traditional dimensionality reduction methods such as principal component analysis (PCA) do not yield satisfactory results for high dimensional data (for example, see\cite{muller2008limitations, jung2009pca, kosztyan2021}). In contrast to these methods, feature screening uses rough but fast techniques to select a larger set that contains most or all of the active predictors\cite{mai2013kolmogorov, mai2015fused, yang2019sufficient}. Although several screening methods have been published for ultrahigh dimensional data in recent years (e.g.,\cite{li2017profile, he2019robust, nandy2021covariate, do2016classifying, do2019latent,roy2022exact}), only a few of them can be used in cases when the response variable contains numerous classes. In particular, in the domains of neuroscience and biometric authentication, datasets with these properties are often encountered.

To reduce various ultrahigh dimensional feature spaces in binary classification problems, Fan and Li (2008)\cite{fan2008sure} proposed a \emph{sure independence screening (SIS)} method in the context of linear regression models. According to Fan and Fan (2008)\cite{fan2008high}, all features that effectively characterize both classes can be extracted by using two-sample $t$-test statistics. Thus resulting in \emph{features annealed independence rules (FAIR)}. For similar binary classification problems, Mai and Zou (2013)\cite{mai2013kolmogorov} used a Kolmogorov filter (KF) method that was extended to handle a multiclass response in Mai and Zou (2013)\cite{mai2015fused}. The KF method is also applied for the ultrahigh dimensional binary classification problem with a dependent variable in Lai et al.~(2017)\cite{lai2017model}. For solving similar tasks, Roy et al.~(2022)\cite{roy2022exact} proposed a model-free feature screening method based on energy distances (see\cite{szekely2005new,baringhaus2010rigid}).

While most existing feature screening approaches are unsuitable for examining higher-order interactive structures and nonlinear structures, random forest (RF)\cite{breiman2001random} can overcome such difficulties\cite{wang2015forest}. To provide a robust screening solution for ultrahigh-dimensional, multiclass data, we propose the \textit{random forest-based multiround screening (RFMS)} method. The Julia package that implements RFMS is publicly available on GitHub\cite{git_rfms}. The RFMS improves the accuracy and scalability of both traditional selection methods and existing RF-based screening by organizing the screening process into rounds. As an advantage, the input is processed in larger chunks, and we can iteratively distill a well-predicting subset of features.

The paper is organized as follows. The Data and methodology section introduces the dataset that was used for benchmarking and the proposed feature screening method. The Results and discussion section presents the performance of the novel screening method and compare it with other reduction algorithms. Finally, the Conclusions and future work section provides our conclusions and suggest future research directions.

\section*{Data and methodology}
\label{sec:data_and_methodology}

\subsection*{Synthetic dataset}
\label{ssec:synthetic_dataset}

\noindent
To compare the performance of the proposed RFMS with a wide range of feature screening methods, an ultrahigh dimensional, multiclass feature space---with ground truth and some additional side information on the usefulness of the features---was employed. This feature space imitates the key properties of the private signature dataset of Cursor Insight---,which was the winner of the ICDAR competition on signature verification and writer identification in 2015\cite{malik2015icdar2015})---. Moreover, it was compiled by using the \textit{BiometricBlender} data generator\cite{stippinger2021blender}. The \textit{BiometricBlender} Python package provides an alternative to real biometric datasets, which are typically not freely accessible and cannot be published for industrial reasons. It is publicly available on GitHub\cite{git_bioblend}.

The following parameters were set during the data generation process:

\begin{itemize}[noitemsep]
\item \texttt{n-classes} = 100;
\item \texttt{n-samples-per-class} = 64;
\item \texttt{n-true-features} = 100;
\item \texttt{n-fake-features} = 300;
\item \texttt{min-usefulness} = 0.5;
\item \texttt{max-usefulness} = 1;
\item \texttt{location-sharing-extent} = 50;
\item \texttt{location-ordering-extent} = 20;
\item \texttt{n-features-out} = 10\,000;
\item \texttt{blending-mode} = `logarithmic';
\item \texttt{min-count} = 4;
\item \texttt{max-count} = 8;
\item \texttt{random-state} = 137.
\end{itemize}

\noindent
The resulting dataset contains a categorical target variable with 100 unique classes and 10\,000 intercorrelated features. Note that neither of the features in itself contains enough information to accurately classify the data over the target variable. However, an appropriate combination can provide sufficient information to achieve classification with high accuracy. Due to the high dimensionality of the dataset, the identification of such a combination is a nontrivial task (regardless of the classification algorithm). The screening algorithm that was introduced in this paper provides a reliable, robust, and resource-efficient means to achieve that goal.

\subsection*{Random forest-based multiround screening}
\label{ssec:multiround_screening}

\noindent
Before we describe the steps of the proposed screening algorithm, several notations must be described. Let $y \in \{1,2,\dots,k\}$ be a categorical target variable that contains $k$ different classes $(k\in\mathbb{N}^+,\, k\geq 2)$, and let $\textbf{x} = \left<x_1,x_2,\dots,x_n\right>$ be the tuple of input features $(n\in\mathbb{N}^+)$. (Note that the method may straightforwardly be applied to continuous target variables as well.) Moreover, let $\alpha, \beta \in \mathbb{N}^+$ be predefined parameters such that $1\leq \beta \leq \alpha \leq n$, where $\alpha$ denotes the size of the subsets that the feature space will be divided into, and $\beta$ denotes the number of features that will be selected by the algorithm. For optimal values of \(\alpha\) and \(\beta\), see the Supplementary information parameters \texttt{step-size} and \texttt{reduced-size}, respectively.

\medskip

\noindent
\textbf{Preparation.} First, the input features of $\textbf{x}$ are arranged in random order. Formally, let $\pi$ be a random permutation of $\{1,2,\dots,n\}$, then
\[
  \textbf{x}_{\pi} = \left<x_{\pi(1)},x_{\pi(2)},\dots,x_{\pi(n)}\right>
\]
denotes the randomly ordered tuple of input features. $\textbf{x}_{\pi}$ is then divided into $m = \lceil n / \alpha \rceil$ subsets as follows:
\begin{equation*}
\begin{split}
  \textbf{x}^1_\pi &= \left<x_{\pi(1)},x_{\pi(2)},\dots,x_{\pi(\alpha)}\right>,\\
  \textbf{x}^2_\pi &= \left<x_{\pi(\alpha+1)},x_{\pi(\alpha+2)},\dots,x_{\pi(2\alpha)}\right>,\\
  \vdots\; &\\
  \textbf{x}^j_\pi &= \left<x_{\pi((j-1)\alpha+1)},x_{\pi((j-1)\alpha+2)},\dots,x_{\pi(j\alpha)}\right> \quad (1\leq j <m),\\
  \vdots\; &\\  
  \textbf{x}^m_\pi &= \left<x_{\pi((m-1)\alpha+1)},x_{\pi((m-1)\alpha+2)},\dots,x_{\pi(n)}\right>.
\end{split}
\end{equation*}

\medskip

\noindent
\textbf{Iteration.} In this step, we iterate over the abovementioned subsets by selecting the $\beta$ most important features from a subset, adding them to the next subset, and repeating this process until the $\beta$ most important features are selected from the last subset. Formally, for $1\leq i \leq m$, let
\[
  \bar{\textbf{x}}^i_\pi = \textbf{x}^i_\pi \frown \textbf{z}^{i-1} = \left<\bar{x}^i_{1},\bar{x}^i_{2},\dots,\bar{x}^i_{t}\right>
\]
(i.e., the concatenation of the two tuples), where $t = |\textbf{x}^i_\pi| + \beta \leq \alpha + \beta$, $\textbf{z}^0 = \left<\,\right>$ is an empty tuple, and $\textbf{z}^i$ $(1\leq i < m)$ will be defined below. (Note that \(\bar{\textbf{x}}^1_\pi = \textbf{x}^1_\pi\).) The relative feature importance of $\bar{\textbf{x}}^i_\pi$ on $y$ is identified by using random forest classification. The importance of a feature is determined by the total number of times it appears in the classification forest (often termed the \emph{selection frequency}).

The most important $\beta$ features of $\bar{\textbf{x}}^i_\pi$ are stored in:
\begin{equation*}
  \textbf{z}^i = \left<\bar{x}^i_{G_i(1)},\bar{x}^i_{G_i(2)},\dots,\bar{x}^i_{G_i(\beta)}\right>,
\end{equation*}

\noindent
where $G_i:\, \{1,2,\dots,\beta\} \rightarrow \{1,2,\dots,t\}$ is an injective function that sorts the features in $\textbf{z}^i$ in descending order of their importance.

\medskip

\noindent
\textbf{Result.} The $\beta$ features considered most important by the RFMS are found in the following:
\begin{equation*}
  \textbf{z} = \textbf{z}^m = \left<\bar{x}^m_{G_m(1)},\bar{x}^m_{G_m(2)},\dots,\bar{x}^m_{G_m(\beta)}\right>.
\end{equation*}

\medskip

\begin{figure}[htbp]
  \centering
  \includegraphics[width=0.85\linewidth]{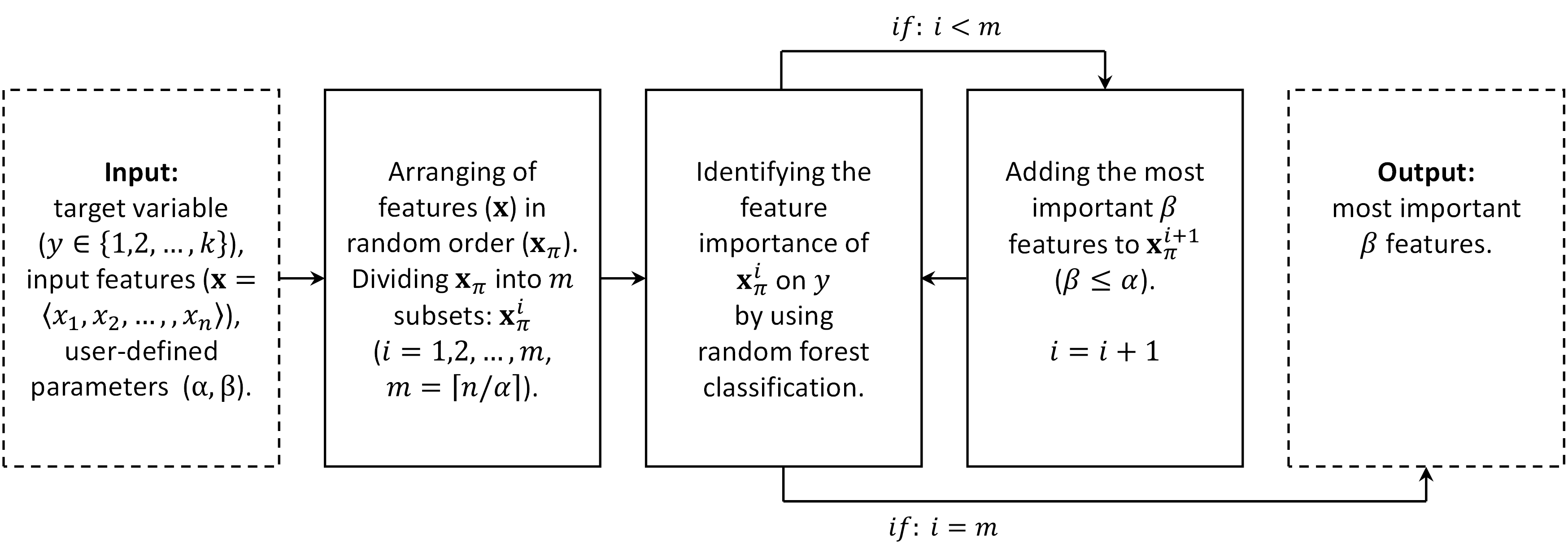}
  \caption{Steps of the RFMS.}\label{fig:flowchart}
\end{figure}

The aforementioned steps of the calculation are illustrated in \Cref{fig:flowchart}.

\section*{Results and discussion}
\label{sec:results}

\begin{samepage}
\noindent
To compare the performance of the RFMS with off-the-shelf screening methods, we completed the following measurements:
\begin{enumerate}
\item We measured the maximum accuracy of three basic classifiers---$k$-nearest neighbors ($k$NN)\cite{fix1989discriminatory,cover1967nearest}, support vector classifier (SVC)\cite{Vapnik1998}, and random forest (RF)\cite{breiman2001random}---on the full feature set by using $n$-fold cross-validation. The optimal parameters of the classifiers were identified via a grid search.
\item We performed screening by using four different methods (including our method), thus resulting in the requested number of screened features (from 10 to 500) per method. The tested screening methods included principal component analysis (PCA)\cite{pearson1901liii,hotelling1933analysis}, factor analysis (FA)\cite{spearman1904general,yong2013beginner}, $k$-best\cite{wong2002vlsi}, and RFMS.
\item We measured the maximum accuracy of the three classifiers on each of the screened feature sets by using \(n\)-fold cross-validation.
\item For every step above, we also measured the CPU usage.
\end{enumerate}
\end{samepage}
\noindent
Note that methods based on neural networks are legally restricted to prevent the restoration of original signatures. Therefore, we did not utilize these methods as a basis for comparison. The highest classification accuracies for each combination, along with their screening and fitting times, are summarized in \Cref{tab:measurements}. The optimized hyperparameters that were used during the application of the RFMS method can be found in the Supplementary information.

\begin{table}[!htb]
    \centering
    \begin{subtable}{.495\linewidth}
      \centering
      \caption{Classification accuracy}
      \resizebox{\textwidth}{!}{%
      \begin{tabular}{llrrrrr}
        \toprule
        \multicolumn{2}{r}{Reduction:} & \multicolumn{1}{c}{None} &  \multicolumn{1}{c}{PCA}        & \multicolumn{1}{c}{FA}   &  \multicolumn{1}{c}{$k$-best}   & \multicolumn{1}{c}{RFMS} \\
        \midrule
        \multirow{3}{*}{\rotatebox{90}{Class.:}}
        & $k$NN & 0.043 & 0.092 & 0.101      & 0.313 & \bf{0.381} \\
        & SVC   & 0.244 & 0.226 & 0.420      & 0.518 & \bf{0.614} \\
        & RF    & 0.428 & 0.155 & \bf{0.614} & 0.533 & 0.604 \\
        \bottomrule
      \end{tabular}
      }
    \end{subtable}%
    \begin{subtable}{.490\linewidth}
      \centering
        \caption{Screening time}
        \resizebox{\textwidth}{!}{%
        \begin{tabular}{llrrrrr}
          \toprule
          \multicolumn{2}{r}{Reduction:} & \multicolumn{1}{c}{None} &  \multicolumn{1}{c}{PCA}        & \multicolumn{1}{c}{FA}   &  \multicolumn{1}{c}{$k$-best}   & \multicolumn{1}{c}{RFMS} \\
          \midrule
          \multirow{3}{*}{\rotatebox{90}{Class.:}}
          & $k$NN & - &  6.5s &  21s & 1.63s & 11\,464s \\
          & SVC   & - & 54.6s & 457s & 1.48s & 11\,266s \\
          & RF    & - & 25.8s & 471s & 1.48s & 10\,931s \\
          \bottomrule
        \end{tabular}
        }
    \end{subtable}
    
    \medskip
    
    \begin{subtable}{.545\linewidth}
      \centering
        \caption{Fitting time}
        \resizebox{\textwidth}{!}{%
        \begin{tabular}{llrrrrr}
          \toprule
          \multicolumn{2}{r}{Reduction:} & \multicolumn{1}{c}{None} &  \multicolumn{1}{c}{PCA}        & \multicolumn{1}{c}{FA}   &  \multicolumn{1}{c}{$k$-best}   & \multicolumn{1}{c}{RFMS} \\
          \midrule
          \multirow{3}{*}{\rotatebox{90}{Class.:}}
          & $k$NN &   0.76s & 0.0062s & 0.0061s & 0.010s & 0.0082s \\
          & SVC   &    185s &   11.9s &   11.9s &  11.3s &   11.6s \\
          & RF    & 1\,145s &    223s &    233s &   192s &   59.6s \\
          \bottomrule
        \end{tabular}
        }
    \end{subtable}
    
    \caption{Classification results on the 6\,400$\times$10\,000 dataset for three basic classifiers and various reduction algorithms. \emph{(a)} Only the best accuracy among all of the parameters is reported. \emph{(b)} Screening times are the CPU times of the feature screening step and correspond to the best accuracy shown above. \emph{(c)} Fitting times are defined as the CPU times after the reduction step and correspond to the best accuracy shown above.}
\label{tab:measurements}
\end{table}

Based on the results, the RFMS and FA methods outperformed both PCA and $k$-best screening in accuracy. The highest accuracy was achieved by using the RFMS--SVC and FA--RF pairs (61.4\%); however, the latter combination required considerably lower screening time. Notably, depending on the persistence of the features (see, e.g.,\cite{friedman2017method}), the screening was performed relatively infrequently in comparison with the fitting procedure, in which the combination comprising RFMS proved to be relatively fast. Furthermore, in exchange for a slower screening procedure, RFMS offers several advantages over the FA method. These advantages are detailed below.

\vskip 1em
\noindent
\textbf{Potential cost reduction in feature computation.} To use FA on an incoming sample, its full feature set must be computed before the transformation can be applied. The trained model only works on the transformed feature set. In contrast, the output of RFMS is a transformation-free subset of the original feature set. This facilitates the interpretation of the resulting features; in addition, once RFMS has finished, and we have the set of optimal features, only these features need to be computed on any further incoming samples. This could be a significant factor in saving on cost and time in a production system.

\vskip 1em
\noindent
\textbf{Suitability for several classifiers.} Although the combination of FA and RF resulted in a high accuracy and low screening time, the accuracy of the same FA output with SVC and \(k\)NN classifiers produced significantly weaker results (accuracy of 42\% and 10\%, respectively). However, for the RFMS output, SVC performed slightly better than RF (just as well as the FA--RF combination), and even the accuracy of the \(k\)NN classifier at 38.1\% was much closer to the top performers.

\vskip 1em
\noindent
\textbf{Robustness.} If we further investigate past the highest accuracies for every combination and observe how the accuracy changes with the adjustment of the hyperparameters, we can conclude that FA is quite sensitive. If we reduce the number of screened features (components) from 500 to 250, the highest achievable accuracy drops to 33.1\%. A further reduction to 125 results in an accuracy of only 25\%. A similar performance drop is observable if we begin to increase the number of features from 500. However, with RFMS, a reduction in the number of screened features from 500 to 200 only slightly reduces the best accuracy to 60.8\%, and with a further reduction to 100, the accuracy is still 55.4\%. We observed this behavior with high probability when the degrees of freedom of the data were well defined, but the FA was requested to produce fewer features.

\begin{figure}[htbp]
  \centering
  \includegraphics[width=0.75\linewidth]{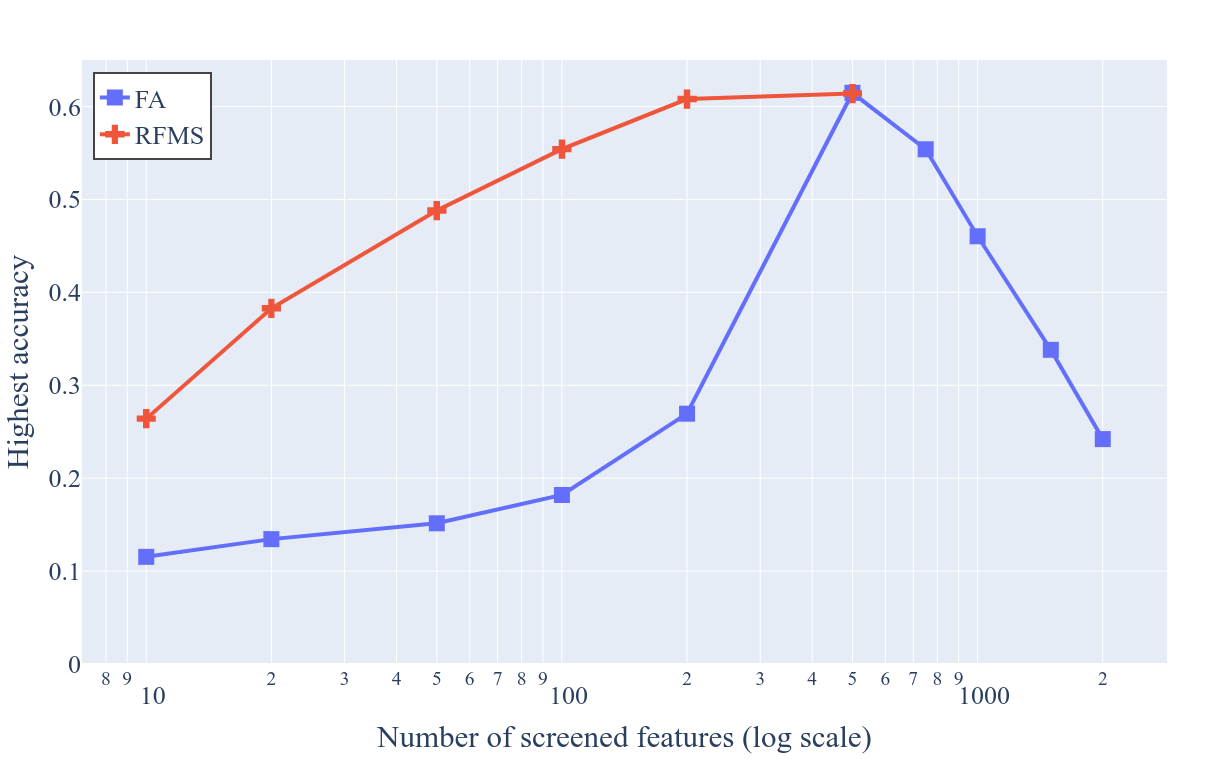}
  \caption{Convergence of highest accuracy as a function of the number of screened features (components). Accuracy is measured on a scale of 0-1. RFMS converges to the optimum much quicker than FA.}\label{fig:convergence}
\end{figure}

\Cref{fig:convergence} summarizes both trends on a single plot, thus demonstrating how the highest achievable accuracy converges to its global optimum as the number of screened features increases. Note that the deviation from the plotted accuracy values with the randomization of the selection and measurement process is negligible.

In addition, by adjusting the RFMS hyperparameters, the screening time can be significantly reduced without compromising the classification accuracy. For example, with the right combination, the screening time can be decreased to 2\,143 s (merely 1/5$^{\mathrm{th}}$ of the highest value in \Cref{tab:measurements}), while the achievable accuracy is still 60\%. The fastest run in our test occurred for 1\,738 s (15\% of the longest screening time), and even that output could achieve a 57.3\% accuracy (93.4\% of the overall highest accuracy).

\vskip 1em
\noindent
\textbf{Performance on proprietary datasets.} We have extensively used RFMS on our own proprietary biometric feature sets; although we cannot publicly share these datasets, we can share our experiences. We found that the FA--RF pair typically performs worse than the combination of RFMS--RF for real feature sets. In one particular case, we trained both screening methods on a dataset of 10\,000 classes, 81\,000 samples, and 18\,700 input features and targeted 200 output features. We subsequently measured the performance of the screened features by using a disjunct dataset of 44 classes and 58\,000 samples (as well as the same number of features). The best classification accuracy that we could obtain on an FA transformed feature set was approximately 82\%, while the RFMS-filtered output could elicit classification rates up to as high as 93\%, albeit with the screening time being significantly longer (both values have been measured with 5-fold cross-validation). However, given the sensitive and proprietary nature of the dataset, we cannot provide hard evidence for this claim.

\section*{Conclusions and future work}
\label{sec:conclusions}

\noindent
Research on feature screening has grown rapidly in recent years; however, screening ultralarge, multiclass data is still in its infancy. To narrow this gap in the research, we presented a novel method known as random forest-based multiround screening (RFMS) that can be effectively applied in such circumstances. Due to the fact that ultrahigh-dimensional, multiclass data are typically encountered in biometrics, the RFMS was benchmarked on a synthetic feature space that imitates the key properties of a real (private) signature dataset. Based on the results, the RFMS is on par with industry-standard feature screening methods, and it also possesses many advantages over these methods due to its flexibility and robustness, as well as its transformation-free operation. The Julia package that implements RFMS is publicly available on GitHub\cite{git_rfms}. 

The difference in maximum accuracy that was achieved on real and synthetic data suggests that the synthetic data generator used for tests does not yet reproduce all of the properties of real data that challenge feature screeners, and this scenario is especially true for factor analysis. Therefore, it would be important to explore the properties of real data that cause this difference and to further develop \emph{BiometricBlender} in this direction, which could subsequently enable more realistic tests.

To further develop the RFMS method, the following future works are suggested:
\begin{enumerate}
\item Filter highly correlated variables in every iteration just before classification, as this could improve the importance of the features that are proposed by the method.
\item Identify the means of automatically determining the number of important features to be retained per cycle, thus allowing for all of the important features to be kept and most unnecessary features to be dropped. This could improve both accuracy and computation time.
\item Reduce screening time by using more parallel computations (random forest building already utilizes multiple threads when available).
\item Replace random forest and the importance metrics with other less common (but potentially better performing) alternatives.
\item Hyperparameter optimization is typically not viable with brute force due to lengthy computation times. Handy visualization tools could provide useful hints for manual boosting.
\item Consider various types of elimination tournaments, such as going through the input several times or using alternative scoring like the Elo or Glicko\cite{glickman1995glicko} systems. This may further improve accuracy when some information is nontrivially distributed in multiple ``entangled'' features.
\end{enumerate}

\appendix
\section*{Supplementary information}

\renewcommand{\thetable}{S\arabic{table}}
\setcounter{table}{0}

\noindent
RFMS was based on a Julia package that is publicly available on GitHub\cite{git_rfms}. Its hyperparameters have been optimized via a grid search to identify a combination that produces the highest classification accuracy, as well as to observe the effect of changing the hyperparameters on the outcome.

In all cases, a fixed random seed of 20\,230\,125 was used to make the process deterministic. A fixed value of 0.7 was set for the \texttt{partial-sampling} parameter. Finally, 100 random features were added to the mix before screening as \emph{canaries}. If any of these features had appeared in the final set of screened features on the output, we could have been confident that any less important features were simply noise. However, none of our total 3\,969 measurements (539 screening configurations combined with 9 different classifiers, minus the contradicting combinations) stumbled upon a random feature among the screened ones; therefore, we were confident that the screening process identified truly relevant and meaningful features.

\Cref{tab:rfms-params} summarizes the best four hyperparameter combinations, one for each of the three tested classifiers, plus one that produced the smallest screening time.

\begin{table}[htb]
  \centering
  \resizebox{0.6\textwidth}{!}{%
  \begin{tabular}{lrrrr}
    \toprule
    Parameter & \multicolumn{1}{c}{\(k\)NN} & \multicolumn{1}{c}{SVC} & \multicolumn{1}{c}{RF} & \multicolumn{1}{c}{Fastest} \\
    \midrule
    \texttt{reduced-size}        &    200 &    500 &    200 &    200 \\
    \texttt{step-size}           &    505 &    505 &    505 & 2\,020 \\
    \texttt{n-subfeatures}       &    200 &    100 &    500 & 1\,000 \\
    \texttt{n-trees}             &    500 & 1\,000 &    200 &    100 \\
    \texttt{min-samples-leaf}    &      1 &      1 &      1 &     40 \\
    \texttt{min-purity-increase} &   0.01 &    0.1 &    0.1 &      0 \\
    \midrule
    Screening time:       & 11\,464s & 10\,931s & 11\,266s & 1\,738s \\
    Accuracy: \\
    \hspace{3em} - with \(k\)NN  & 38.1\% &        &        & 33.9\% \\
    \hspace{3em} - with SVC      &        & 61.4\% &        & 56.7\% \\
    \hspace{3em} - with RF       &        &        & 60.4\% & 57.3\% \\
    \bottomrule
  \end{tabular}
  }
  \caption{Optimal screening hyperparameters, the corresponding screening times, and the best achievable classification accuracies for various classifiers and fastest screening according to grid search results.}
  \label{tab:rfms-params}
\end{table}

\section*{Data availability}

The applied feature space was compiled by using the \textit{BiometricBlender} data generator\cite{stippinger2021blender}, which is publicly available on GitHub\cite{git_bioblend}.

\bibliography{bibliography}

\begin{thebibliography}{10}
\urlstyle{rm}
\expandafter\ifx\csname url\endcsname\relax
  \def\url#1{\texttt{#1}}\fi
\expandafter\ifx\csname urlprefix\endcsname\relax\def\urlprefix{URL }\fi
\expandafter\ifx\csname doiprefix\endcsname\relax\def\doiprefix{DOI: }\fi
\providecommand{\bibinfo}[2]{#2}
\providecommand{\eprint}[2][]{\url{#2}}

\bibitem{wang2009forward}
\bibinfo{author}{Wang, H.}
\newblock \bibinfo{journal}{\bibinfo{title}{Forward regression for ultra-high
  dimensional variable screening}}.
\newblock {\emph{\JournalTitle{Journal of the American Statistical
  Association}}} \textbf{\bibinfo{volume}{104}}, \bibinfo{pages}{1512--1524},
  \doiprefix\url{10.1198/jasa.2008.tm08516} (\bibinfo{year}{2009}).

\bibitem{tan2022feature}
\bibinfo{author}{Tan, H.}, \bibinfo{author}{Wang, G.}, \bibinfo{author}{Wang,
  W.} \& \bibinfo{author}{Zhang, Z.}
\newblock \bibinfo{journal}{\bibinfo{title}{Feature selection based on distance
  correlation: a filter algorithm}}.
\newblock {\emph{\JournalTitle{Journal of Applied Statistics}}}
  \textbf{\bibinfo{volume}{49}}, \bibinfo{pages}{411--426},
  \doiprefix\url{10.1080/02664763.2020.1815672} (\bibinfo{year}{2022}).

\bibitem{clarke2008properties}
\bibinfo{author}{Clarke, R.} \emph{et~al.}
\newblock \bibinfo{journal}{\bibinfo{title}{The properties of high-dimensional
  data spaces: implications for exploring gene and protein expression data}}.
\newblock {\emph{\JournalTitle{Nature reviews cancer}}}
  \textbf{\bibinfo{volume}{8}}, \bibinfo{pages}{37--49},
  \doiprefix\url{10.1038/nrc2294} (\bibinfo{year}{2008}).

\bibitem{johnstone2009statistical}
\bibinfo{author}{Johnstone, I.~M.} \& \bibinfo{author}{Titterington, D.~M.}
\newblock \bibinfo{title}{Statistical challenges of high-dimensional data},
  \doiprefix\url{10.1098/rsta.2009.0159} (\bibinfo{year}{2009}).

\bibitem{li2017challenges}
\bibinfo{author}{Li, J.} \& \bibinfo{author}{Liu, H.}
\newblock \bibinfo{journal}{\bibinfo{title}{Challenges of feature selection for
  big data analytics}}.
\newblock {\emph{\JournalTitle{IEEE Intelligent Systems}}}
  \textbf{\bibinfo{volume}{32}}, \bibinfo{pages}{9--15},
  \doiprefix\url{10.1109/MIS.2017.38} (\bibinfo{year}{2017}).

\bibitem{ferri1994}
\bibinfo{author}{Ferri, F.~J.}, \bibinfo{author}{Pudil, P.},
  \bibinfo{author}{Hatef, M.} \& \bibinfo{author}{Kittler, J.}
\newblock \bibinfo{title}{Comparative study of techniques for large-scale
  feature selection}.
\newblock In \emph{\bibinfo{booktitle}{Machine Intelligence and Pattern
  Recognition}}, vol.~\bibinfo{volume}{16}, \bibinfo{pages}{403--413},
  \doiprefix\url{10.1016/B978-0-444-81892-8.50040-7}
  (\bibinfo{publisher}{Elsevier}, \bibinfo{year}{1994}).

\bibitem{ni2017adjusted}
\bibinfo{author}{Ni, L.}, \bibinfo{author}{Fang, F.} \& \bibinfo{author}{Wan,
  F.}
\newblock \bibinfo{journal}{\bibinfo{title}{Adjusted {Pearson Chi-Square}
  feature screening for multi-classification with ultrahigh dimensional data}}.
\newblock {\emph{\JournalTitle{Metrika}}} \textbf{\bibinfo{volume}{80}},
  \bibinfo{pages}{805--828}, \doiprefix\url{10.1007/s00184-017-0629-9}
  (\bibinfo{year}{2017}).

\bibitem{borutaalg}
\bibinfo{author}{Kursa, M.~B.} \& \bibinfo{author}{Rudnicki, W.~R.}
\newblock \bibinfo{journal}{\bibinfo{title}{Feature selection with the {Boruta
  Package}}}.
\newblock {\emph{\JournalTitle{Journal of Statistical Software}}}
  \textbf{\bibinfo{volume}{36}}, \bibinfo{pages}{1–13},
  \doiprefix\url{10.18637/jss.v036.i11} (\bibinfo{year}{2010}).

\bibitem{speiser2019}
\bibinfo{author}{Speiser, J.~L.}, \bibinfo{author}{Miller, M.~E.},
  \bibinfo{author}{Tooze, J.} \& \bibinfo{author}{Ip, E.}
\newblock \bibinfo{journal}{\bibinfo{title}{A comparison of random forest
  variable selection methods for classification prediction modeling}}.
\newblock {\emph{\JournalTitle{Expert Systems with Applications}}}
  \textbf{\bibinfo{volume}{134}}, \bibinfo{pages}{93--101},
  \doiprefix\url{10.1016/j.eswa.2019.05.028} (\bibinfo{year}{2019}).

\bibitem{muller2008limitations}
\bibinfo{author}{Muller, K.~E.}, \bibinfo{author}{Chi, Y.-Y.},
  \bibinfo{author}{Ahn, J.} \& \bibinfo{author}{Marron, J.}
\newblock \bibinfo{journal}{\bibinfo{title}{Limitations of high dimension, low
  sample size principal components for {Gaussian} data}}.
\newblock {\emph{\JournalTitle{under revision for resubmission}}}
  (\bibinfo{year}{2008}).

\bibitem{jung2009pca}
\bibinfo{author}{Jung, S.} \& \bibinfo{author}{Marron, J.~S.}
\newblock \bibinfo{journal}{\bibinfo{title}{{PCA} consistency in high
  dimension, low sample size context}}.
\newblock {\emph{\JournalTitle{The Annals of Statistics}}}
  \textbf{\bibinfo{volume}{37}}, \bibinfo{pages}{4104--4130},
  \doiprefix\url{10.1214/09-AOS709} (\bibinfo{year}{2009}).

\bibitem{kosztyan2021}
\bibinfo{author}{Koszty{\'a}n, Z.~T.}, \bibinfo{author}{Kurbucz, M.~T.} \&
  \bibinfo{author}{Katona, A.~I.}
\newblock \bibinfo{journal}{\bibinfo{title}{Network-based dimensionality
  reduction of high-dimensional, low-sample-size datasets}}.
\newblock {\emph{\JournalTitle{Knowledge-Based Systems}}}
  \bibinfo{pages}{109180}, \doiprefix\url{10.1016/j.knosys.2022.109180}
  (\bibinfo{year}{2022}).

\bibitem{mai2013kolmogorov}
\bibinfo{author}{Mai, Q.} \& \bibinfo{author}{Zou, H.}
\newblock \bibinfo{journal}{\bibinfo{title}{The {K}olmogorov filter for
  variable screening in high-dimensional binary classification}}.
\newblock {\emph{\JournalTitle{Biometrika}}} \textbf{\bibinfo{volume}{100}},
  \bibinfo{pages}{229--234}, \doiprefix\url{10.1093/biomet/ass062}
  (\bibinfo{year}{2013}).

\bibitem{mai2015fused}
\bibinfo{author}{Mai, Q.}, \bibinfo{author}{Zou, H.} \emph{et~al.}
\newblock \bibinfo{journal}{\bibinfo{title}{The fused {K}olmogorov filter: {A}
  nonparametric model-free screening method}}.
\newblock {\emph{\JournalTitle{The Annals of Statistics}}}
  \textbf{\bibinfo{volume}{43}}, \bibinfo{pages}{1471--1497},
  \doiprefix\url{10.1214/14-AOS1303} (\bibinfo{year}{2015}).

\bibitem{yang2019sufficient}
\bibinfo{author}{Yang, B.}, \bibinfo{author}{Yin, X.} \&
  \bibinfo{author}{Zhang, N.}
\newblock \bibinfo{journal}{\bibinfo{title}{Sufficient variable selection using
  independence measures for continuous response}}.
\newblock {\emph{\JournalTitle{Journal of Multivariate Analysis}}}
  \textbf{\bibinfo{volume}{173}}, \bibinfo{pages}{480--493},
  \doiprefix\url{10.1016/j.jmva.2019.04.006} (\bibinfo{year}{2019}).

\bibitem{li2017profile}
\bibinfo{author}{Li, Y.}, \bibinfo{author}{Li, G.}, \bibinfo{author}{Lian, H.}
  \& \bibinfo{author}{Tong, T.}
\newblock \bibinfo{journal}{\bibinfo{title}{Profile forward regression
  screening for ultra-high dimensional semiparametric varying coefficient
  partially linear models}}.
\newblock {\emph{\JournalTitle{Journal of Multivariate Analysis}}}
  \textbf{\bibinfo{volume}{155}}, \bibinfo{pages}{133--150},
  \doiprefix\url{10.1016/j.jmva.2016.12.006} (\bibinfo{year}{2017}).

\bibitem{he2019robust}
\bibinfo{author}{He, Y.}, \bibinfo{author}{Zhang, L.}, \bibinfo{author}{Ji, J.}
  \& \bibinfo{author}{Zhang, X.}
\newblock \bibinfo{journal}{\bibinfo{title}{Robust feature screening for
  elliptical copula regression model}}.
\newblock {\emph{\JournalTitle{Journal of Multivariate Analysis}}}
  \textbf{\bibinfo{volume}{173}}, \bibinfo{pages}{568--582},
  \doiprefix\url{10.1016/j.jmva.2019.05.003} (\bibinfo{year}{2019}).

\bibitem{nandy2021covariate}
\bibinfo{author}{Nandy, D.}, \bibinfo{author}{Chiaromonte, F.} \&
  \bibinfo{author}{Li, R.}
\newblock \bibinfo{journal}{\bibinfo{title}{Covariate information number for
  feature screening in ultrahigh-dimensional supervised problems}}.
\newblock {\emph{\JournalTitle{Journal of the American Statistical
  Association}}} \bibinfo{pages}{1--14},
  \doiprefix\url{10.1080/01621459.2020.1864380} (\bibinfo{year}{2021}).

\bibitem{do2016classifying}
\bibinfo{author}{Do, T.-N.} \& \bibinfo{author}{Poulet, F.}
\newblock \bibinfo{title}{Classifying very high-dimensional and large-scale
  multi-class image datasets with {Latent-lSVM}}.
\newblock In \emph{\bibinfo{booktitle}{2016 intl IEEE conferences on ubiquitous
  intelligence \& computing, advanced and trusted computing, scalable computing
  and communications, cloud and big data computing, internet of people, and
  smart world congress (uic/atc/scalcom/cbdcom/iop/smartworld)}},
  \bibinfo{pages}{714--721},
  \doiprefix\url{10.1109/UIC-ATC-ScalCom-CBDCom-IoP-SmartWorld.2016.0116}
  (\bibinfo{organization}{IEEE}, \bibinfo{year}{2016}).

\bibitem{do2019latent}
\bibinfo{author}{Do, T.-N.} \& \bibinfo{author}{Poulet, F.}
\newblock \bibinfo{journal}{\bibinfo{title}{{Latent-lSVM} classification of
  very high-dimensional and large-scale multi-class datasets}}.
\newblock {\emph{\JournalTitle{Concurrency and Computation: Practice and
  Experience}}} \textbf{\bibinfo{volume}{31}}, \bibinfo{pages}{e4224},
  \doiprefix\url{10.1002/cpe.4224} (\bibinfo{year}{2019}).

\bibitem{roy2022exact}
\bibinfo{author}{Roy, S.}, \bibinfo{author}{Sarkar, S.},
  \bibinfo{author}{Dutta, S.} \& \bibinfo{author}{Ghosh, A.~K.}
\newblock \bibinfo{journal}{\bibinfo{title}{On exact feature screening in
  ultrahigh-dimensional binary classification}}.
\newblock {\emph{\JournalTitle{arXiv preprint arXiv:2205.03831}}}
  \doiprefix\url{10.48550/arXiv.2205.03831} (\bibinfo{year}{2022}).

\bibitem{fan2008sure}
\bibinfo{author}{Fan, J.} \& \bibinfo{author}{Lv, J.}
\newblock \bibinfo{journal}{\bibinfo{title}{Sure independence screening for
  ultrahigh dimensional feature space}}.
\newblock {\emph{\JournalTitle{Journal of the Royal Statistical Society: Series
  B (Statistical Methodology)}}} \textbf{\bibinfo{volume}{70}},
  \bibinfo{pages}{849--911}, \doiprefix\url{10.1111/j.1467-9868.2008.00674.x}
  (\bibinfo{year}{2008}).

\bibitem{fan2008high}
\bibinfo{author}{Fan, J.} \& \bibinfo{author}{Fan, Y.}
\newblock \bibinfo{journal}{\bibinfo{title}{High dimensional classification
  using features annealed independence rules}}.
\newblock {\emph{\JournalTitle{Annals of statistics}}}
  \textbf{\bibinfo{volume}{36}}, \bibinfo{pages}{2605},
  \doiprefix\url{10.1214/07-AOS504} (\bibinfo{year}{2008}).

\bibitem{lai2017model}
\bibinfo{author}{Lai, P.}, \bibinfo{author}{Song, F.}, \bibinfo{author}{Chen,
  K.} \& \bibinfo{author}{Liu, Z.}
\newblock \bibinfo{journal}{\bibinfo{title}{Model free feature screening with
  dependent variable in ultrahigh dimensional binary classification}}.
\newblock {\emph{\JournalTitle{Statistics \& Probability Letters}}}
  \textbf{\bibinfo{volume}{125}}, \bibinfo{pages}{141--148},
  \doiprefix\url{10.1016/j.spl.2017.02.011} (\bibinfo{year}{2017}).

\bibitem{szekely2005new}
\bibinfo{author}{Sz{\'e}kely, G.~J.} \& \bibinfo{author}{Rizzo, M.~L.}
\newblock \bibinfo{journal}{\bibinfo{title}{A new test for multivariate
  normality}}.
\newblock {\emph{\JournalTitle{Journal of Multivariate Analysis}}}
  \textbf{\bibinfo{volume}{93}}, \bibinfo{pages}{58--80},
  \doiprefix\url{10.1016/j.jmva.2003.12.002} (\bibinfo{year}{2005}).

\bibitem{baringhaus2010rigid}
\bibinfo{author}{Baringhaus, L.} \& \bibinfo{author}{Franz, C.}
\newblock \bibinfo{journal}{\bibinfo{title}{Rigid motion invariant two-sample
  tests}}.
\newblock {\emph{\JournalTitle{Statistica Sinica}}} \bibinfo{pages}{1333--1361}
  (\bibinfo{year}{2010}).

\bibitem{breiman2001random}
\bibinfo{author}{Breiman, L.}
\newblock \bibinfo{journal}{\bibinfo{title}{Random forests}}.
\newblock {\emph{\JournalTitle{Machine learning}}}
  \textbf{\bibinfo{volume}{45}}, \bibinfo{pages}{5--32},
  \doiprefix\url{10.1023/A:101093340432} (\bibinfo{year}{2001}).

\bibitem{wang2015forest}
\bibinfo{author}{Wang, G.}, \bibinfo{author}{Fu, G.} \&
  \bibinfo{author}{Corcoran, C.}
\newblock \bibinfo{journal}{\bibinfo{title}{A forest-based feature screening
  approach for large-scale genome data with complex structures}}.
\newblock {\emph{\JournalTitle{BMC genetics}}} \textbf{\bibinfo{volume}{16}},
  \bibinfo{pages}{1--11}, \doiprefix\url{10.1186/s12863-015-0294-9}
  (\bibinfo{year}{2015}).

\bibitem{git_rfms}
\bibinfo{author}{Hancz\'ar, G.} \emph{et~al.}
\newblock \bibinfo{journal}{\bibinfo{title}{{FeatureScreening}}}.
\newblock {\emph{\JournalTitle{GitHub}}}  (\bibinfo{year}{2023}).

\bibitem{malik2015icdar2015}
\bibinfo{author}{Malik, M.~I.} \emph{et~al.}
\newblock \bibinfo{title}{{ICDAR2015 competition on signature verification and
  writer identification for on- and off-line skilled forgeries
  (SigWIcomp2015)}}.
\newblock In \emph{\bibinfo{booktitle}{2015 13th International Conference on
  Document Analysis and Recognition (ICDAR)}}, \bibinfo{pages}{1186--1190},
  \doiprefix\url{10.1109/ICDAR.2015.7333948} (\bibinfo{organization}{IEEE},
  \bibinfo{year}{2015}).

\bibitem{stippinger2021blender}
\bibinfo{author}{Stippinger, M.} \emph{et~al.}
\newblock \bibinfo{journal}{\bibinfo{title}{Biometricblender: Ultra-high
  dimensional, multi-class synthetic data generator to imitate biometric
  feature space}}.
\newblock {\emph{\JournalTitle{SoftwareX}}} \textbf{\bibinfo{volume}{22}},
  \bibinfo{pages}{101366} (\bibinfo{year}{2023}).

\bibitem{git_bioblend}
\bibinfo{author}{Stippinger, M.} \emph{et~al.}
\newblock \bibinfo{journal}{\bibinfo{title}{{BiometricBlender}}}.
\newblock {\emph{\JournalTitle{GitHub}}}  (\bibinfo{year}{2022}).

\bibitem{fix1989discriminatory}
\bibinfo{author}{Fix, E.} \& \bibinfo{author}{Hodges, J.~L.}
\newblock \bibinfo{journal}{\bibinfo{title}{Discriminatory analysis.
  nonparametric discrimination: Consistency properties}}.
\newblock {\emph{\JournalTitle{International Statistical Review / Revue
  Internationale de Statistique}}} \textbf{\bibinfo{volume}{57}},
  \bibinfo{pages}{238--247}, \doiprefix\url{10.2307/1403797}
  (\bibinfo{year}{1989}).

\bibitem{cover1967nearest}
\bibinfo{author}{Cover, T.} \& \bibinfo{author}{Hart, P.}
\newblock \bibinfo{journal}{\bibinfo{title}{Nearest neighbor pattern
  classification}}.
\newblock {\emph{\JournalTitle{IEEE transactions on information theory}}}
  \textbf{\bibinfo{volume}{13}}, \bibinfo{pages}{21--27},
  \doiprefix\url{10.1109/TIT.1967.1053964} (\bibinfo{year}{1967}).

\bibitem{Vapnik1998}
\bibinfo{author}{Vapnik, V.~N.}
\newblock \emph{\bibinfo{title}{Statistical Learning Theory}}
  (\bibinfo{publisher}{Wiley-Interscience}, \bibinfo{year}{1998}).

\bibitem{pearson1901liii}
\bibinfo{author}{Pearson, K.}
\newblock \bibinfo{journal}{\bibinfo{title}{Liii. on lines and planes of
  closest fit to systems of points in space}}.
\newblock {\emph{\JournalTitle{The London, Edinburgh, and Dublin philosophical
  magazine and journal of science}}} \textbf{\bibinfo{volume}{2}},
  \bibinfo{pages}{559--572}, \doiprefix\url{10.1080/14786440109462720}
  (\bibinfo{year}{1901}).

\bibitem{hotelling1933analysis}
\bibinfo{author}{Hotelling, H.}
\newblock \bibinfo{journal}{\bibinfo{title}{Analysis of a complex of
  statistical variables into principal components.}}
\newblock {\emph{\JournalTitle{Journal of educational psychology}}}
  \textbf{\bibinfo{volume}{24}}, \bibinfo{pages}{417},
  \doiprefix\url{10.1037/h0070888} (\bibinfo{year}{1933}).

\bibitem{spearman1904general}
\bibinfo{author}{Spearman, C.}
\newblock \bibinfo{journal}{\bibinfo{title}{{``General Intelligence,''
  Objectively Determined and Measured}}}.
\newblock {\emph{\JournalTitle{The American Journal of Psychology}}}
  \textbf{\bibinfo{volume}{15}}, \bibinfo{pages}{201--292},
  \doiprefix\url{10.2307/1412107} (\bibinfo{year}{1904}).

\bibitem{yong2013beginner}
\bibinfo{author}{Yong, A.~G.}, \bibinfo{author}{Pearce, S.} \emph{et~al.}
\newblock \bibinfo{journal}{\bibinfo{title}{A beginner’s guide to factor
  analysis: Focusing on exploratory factor analysis}}.
\newblock {\emph{\JournalTitle{Tutorials in quantitative methods for
  psychology}}} \textbf{\bibinfo{volume}{9}}, \bibinfo{pages}{79--94},
  \doiprefix\url{10.20982/tqmp.09.2.p079} (\bibinfo{year}{2013}).

\bibitem{wong2002vlsi}
\bibinfo{author}{Wong, K.-w.}, \bibinfo{author}{Tsui, C.-y.},
  \bibinfo{author}{Cheng, R.-K.} \& \bibinfo{author}{Mow, W.-h.}
\newblock \bibinfo{title}{A vlsi architecture of a k-best lattice decoding
  algorithm for mimo channels}.
\newblock In \emph{\bibinfo{booktitle}{2002 IEEE International Symposium on
  Circuits and Systems (ISCAS)}}, vol.~\bibinfo{volume}{3},
  \bibinfo{pages}{III--III}, \doiprefix\url{10.1109/ISCAS.2002.1010213}
  (\bibinfo{organization}{IEEE}, \bibinfo{year}{2002}).

\bibitem{friedman2017method}
\bibinfo{author}{Friedman, L.}, \bibinfo{author}{Nixon, M.~S.} \&
  \bibinfo{author}{Komogortsev, O.~V.}
\newblock \bibinfo{journal}{\bibinfo{title}{Method to assess the temporal
  persistence of potential biometric features: Application to oculomotor, gait,
  face and brain structure databases}}.
\newblock {\emph{\JournalTitle{PloS one}}} \textbf{\bibinfo{volume}{12}},
  \bibinfo{pages}{e0178501}, \doiprefix\url{10.1371/journal.pone.0178501}
  (\bibinfo{year}{2017}).

\bibitem{glickman1995glicko}
\bibinfo{author}{Glickman, M.~E.}
\newblock \bibinfo{journal}{\bibinfo{title}{The glicko system}}.
\newblock {\emph{\JournalTitle{Boston University}}}
  \textbf{\bibinfo{volume}{16}}, \bibinfo{pages}{16--17}
  (\bibinfo{year}{1995}).

\bibitem{Heder2022}
\bibinfo{author}{H{\'{e}}der, M.} \emph{et~al.}
\newblock \bibinfo{journal}{\bibinfo{title}{The past, present and future of the
  {ELKH} cloud}}.
\newblock {\emph{\JournalTitle{Inform{\'{a}}ci{\'{o}}s T{\'{a}}rsadalom}}}
  \textbf{\bibinfo{volume}{22}}, \bibinfo{pages}{128},
  \doiprefix\url{10.22503/inftars.xxii.2022.2.8} (\bibinfo{year}{2022}).

\end{thebibliography}

\section*{Acknowledgments}

The authors would like to thank Erika Griechisch and J\'ulia Bor\'oka N\'emeth (Cursor Insight, London) and Andr\'as Telcs (Wigner Research Centre for Physics, Budapest) for their valuable comments and advice. M.S., M.T.K., and Z.S. thank the support of E\"otv\"os Lor\'and Research Network, grant SA-114/2021 and on behalf of the project ``Identifying Hidden Common Causes: New Data Analysis Methods'' for access to the ELKH Cloud (see\cite{Heder2022}; \url{https://science-cloud.hu/}), which helped us to achieve the results that are published in this paper. Z.S. and M.T.K. received support from the Hungarian Scientific Research Fund (OTKA/NRDI Office) under contract numbers K135837 and PD142593, respectively. M.T.K. thanks the support of the Ministry of Innovation and Technology NRDI Office within the framework of the MILAB Artificial Intelligence National Laboratory Program.

\section*{Author contributions statement}

\textbf{Gergely Hancz\'ar}: Conceptualization, Supervision, Methodology, Writing - Original Draft, Writing - Review \& Editing, Project administration; \textbf{Marcell Stippinger}: Software, Methodology, Validation, Formal analysis, Investigation, Writing - Original Draft, Writing - Review \& Editing; \textbf{D\'avid Han\'ak}:  Software, Methodology, Validation, Formal analysis, Investigation, Visualization, Writing - Original Draft, Writing - Review \& Editing; \textbf{Marcell T. Kurbucz}: Methodology, Investigation, Visualization, Writing - Original Draft, Writing - Review \& Editing; \textbf{Oliv\'er M. T\"orteli}: Software, Validation, Formal analysis, Investigation, Data Curation; \textbf{\'Agnes Chripk\'o}: Formal analysis, Investigation, Writing - Original Draft, Writing - Review \& Editing; \textbf{Zolt\'an Somogyv\'ari}: Supervision, Methodology, Validation, Writing - Original Draft, Writing - Review \& Editing.

\section*{Declaration of competing interests}

\noindent
We wish to make readers aware of the following facts that may be considered potential conflicts of interest, as well as make them aware of significant financial contributions to this work. The nature of the potential conflict of interest involves the fact that some of the authors work for Cursor Insight, which is an IT company targeting human motion analysis, person classification, and identification based on large-scale biometric data in particular.

\section*{Additional information}

The corresponding author is responsible for submitting a \href{http://www.nature.com/srep/policies/index.html#competing}{competing interests statement} on behalf of all authors of the paper.

\end{document}